\documentclass[10pt,twocolumn,letterpaper]{article}

\usepackage{titling}
\usepackage{cvpr}
\usepackage{times}
\usepackage{epsfig}
\usepackage{graphicx}
\usepackage{amsmath}
\usepackage{amssymb}
\usepackage{booktabs}
\usepackage{caption}
\usepackage{subcaption}
\usepackage{color}

\usepackage{stmaryrd} 

\usepackage{fixmath}

\newcommand*\rot{\rotatebox{90}}

\newcommand{\SL}{{\cal L}}

\definecolor{fettrot}{RGB}{255,10,10}

\newcommand{\myparagraph}[1]{\noindent\textbf{#1}}

\usepackage[pagebackref=true,breaklinks=true,letterpaper=true,colorlinks,bookmarks=false]{hyperref}

\cvprfinalcopy 

\def\cvprPaperID{****} 

\ifcvprfinal\fi
\begin{document}

\title{InstanceCut: from Edges to Instances with MultiCut}

\author{Alexander Kirillov$^{1}$\quad Evgeny  Levinkov$^2$ \quad Bjoern  Andres$^2$ \quad Bogdan  Savchynskyy$^1$ \quad Carsten  Rother$^1$ \\
$^1$TU Dresden, Dresden, Germany \quad $^2$MPI for Informatics, Saarbr{\"u}cken, Germany \\
{\tt \small $^1$name.surname@tu-dresden.de} \quad {\tt \small $^2$surname@mpi-inf.mpg.de}
}

\maketitle

\begin{abstract}
This work addresses the task of instance-aware semantic segmentation. Our key motivation is to design a simple method with a new modelling-paradigm, which therefore has a different trade-off between advantages and disadvantages compared to known approaches. 
Our approach, we term InstanceCut, represents the problem by two output modalities: (i) an instance-agnostic semantic segmentation and (ii) all instance-boundaries. The former is computed from a standard convolutional neural network for semantic segmentation, and the latter is derived from a new instance-aware edge detection model. To reason globally about the optimal partitioning of an image
into instances, we combine these two modalities into a novel MultiCut formulation. 
We evaluate our approach on the challenging CityScapes dataset. Despite the conceptual simplicity of our approach, we achieve the best result among all published methods, and perform particularly well for rare object classes.
\end{abstract}

\section{Introduction}
\begin{figure}
	\begin{subfigure}{0.49\linewidth}
		\includegraphics[width=\linewidth]{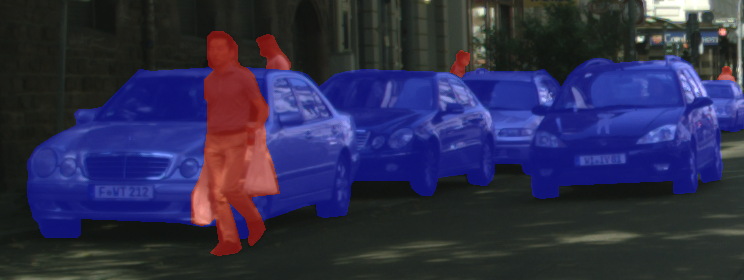}
		\caption{}\label{fig:semantic_gt}
	\end{subfigure}%
	\hspace{0.03cm}
	\begin{subfigure}{0.49\linewidth}
		\includegraphics[width=\linewidth]{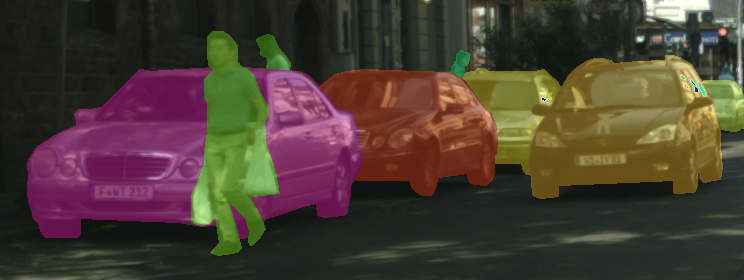}
		\caption{}\label{fig:instance_gt}
	\end{subfigure} \\
	\begin{subfigure}{0.49\linewidth}
		\includegraphics[width=\linewidth]{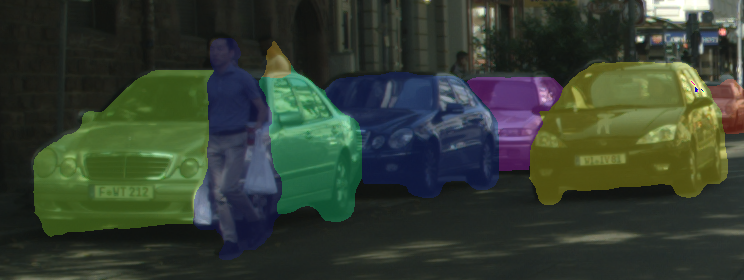}
		\caption{}\label{fig:instance_pred}
	\end{subfigure}%
	\hspace{0.03cm}
	\begin{subfigure}{0.49\linewidth}
		\includegraphics[width=\linewidth]{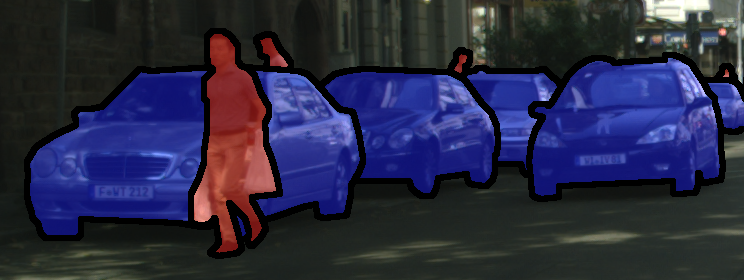}
		\caption{}\label{fig:instance_boundary_gt}
	\end{subfigure}%
	\caption{An image from the CityScapes dataset~\cite{Cordts2016Cityscapes}:
(a) Ground truth semantic segmentation, where all cars have the same label. (b) The ground truth
instance segmentation, where each instance, i.e. object, is highlighted by a distinct color. In this work
we use a ``limiting'' definition of instance segmentation, in the sense that each instance must be a connected component. Despite this limitation, we will demonstrate high-quality results. (c) Shows the result of our InstanceCut method. As can be seen, the front car is split into two instances, in contrast to (b). (d) Our connected-component instances are defined via two output modalities: (i) the semantic segmentation, (ii) all instance-boundaries (shown in bold-black).}
	\label{fig:instance}
\end{figure}

This work addresses the task of segmenting each individual instance of a semantic class in an image. The task is known as {\em instance-aware semantic segmentation}, in short {\em instance segmentation}, and is a more refined task than semantic segmentation, where each pixel is only labeled with its semantic class. An example of semantic segmentation and instance segmentation is shown in Fig.~\ref{fig:semantic_gt}-\ref{fig:instance_gt}.  While semantic segmentation has been a very popular problem to work on in the last half decade, the interest in instance segmentation has significantly increased recently. This is not surprising since semantic segmentation has already reached a high level of accuracy, in contrast to the harder task of instance segmentation. Also, from an application perspective there are many systems, such as autonomous driving or robotics, where a more detailed understanding of the surrounding is important for acting correctly in the world.

In recent years, Convolutional Neural Networks (CNN) have tremendously increased the performance
of many computer vision tasks. This is also true for the task of instance segmentation, see the benchmarks~\cite{Cordts2016Cityscapes,lin2014microsoft}. However, for this task it is, in our view, not clear whether the best modelling-paradigm has
already been found. Hence, the motivation of this work is to explore a new, and very different,
modelling-paradigm. To be more precise, we believe that the problem of instance segmentation has
four core challenges, which any method has to address. Firstly, the label of an instance, e.g. ``car
number 5'', does not have a meaning, in contrast to semantic segmentation, e.g. class ``cars''.
Secondly, the number of instances in an image can vary greatly, e.g. between $0$ and $120$ for an
image in the CityScapes dataset \cite{Cordts2016Cityscapes}. Thirdly, in contrast to object detection
with bounding boxes, each instance (a bounding box) cannot simply be described by four numbers
(corners of bounding box), but has to be described by a set of pixels. Finally, in contrast to semantic
segmentation, a more refined labeling of the training data is needed, i.e. each instance has to be
segmented separately. Especially for rare classes, e.g. motorcycles, the amount of training data,
which is available nowadays, may not be sufficient.
Despite these challenges, the state of the art techniques for instance segmentation are CNN-based. As an example, ~\cite{dai2015instance,zagoruyko2016multipath} address these challenges with a complex multi-loss cascade CNN architectures, which
are, however, difficult to train. In contrast, our modelling-paradigm is very different to standard CNN-based architectures: assume that each pixel is assigned to one semantic class, and additionally we
insert some edges (in-between pixels) which form loops -- then we have solved the problem of
instance segmentation! Each connected region, enclosed by a loop of instance-aware edges is an
individual instance where the class labels of the interior pixels define its class. These are exactly the ingredients of our approach: (i) a standard CNN that outputs an instance-agnostic semantic
segmentation, and (ii) a new CNN that outputs all boundaries of instances. In order to make sure that
instance-boundaries encircle a connected component, and that the interior of a component has the
same class label, we combine these two outputs into a novel multi-cut formulation. We call our approach {\it InstanceCut}.

Our InstanceCut approach has some advantages and disadvantages, which we discuss next. With respect to this, we
would like to stress that these pros and cons are, however, quite different to existing approaches.
This means that in the future we envision that our approach may play an important role, as a subcomponent in an
``ultimate'' instance segmentation system. Let us first consider the limitations, and then the advantages. The minor limitation of our approach is that, obviously, we
cannot find instances that are formed by disconnected regions in the image (see Fig.~\ref{fig:instance_gt}-\ref{fig:instance_pred}). 
However, despite this limitation, we demonstrate results that exceed all
published methods in terms of accuracy. In the future, we foresee various ways to overcome this
limitation, e.g. by reasoning globally about shape.

We see the following major advantages of our approach. Firstly, all the four
challenges for instance segmentation methods, listed above, are addressed in an elegant way: (i) the
multi-cut formulation does not need a unique label for an instance; (ii) the number of instances
arises naturally from the solution of the multi-cut; (iii) our formulation is on the pixel (superpixel)
level; (iv) since we do not train a CNN for segmenting instances globally, our approach deals very well
with instances of rare classes, as they do not need special treatment. Finally, our InstanceCut
approach has another major advantage, from a practical perspective. We can employ any semantic
segmentation method, as long as it provides pixel-wise log-probabilities for each class. Therefore,
advances in this field may directly translate to an improvement of our method. Also, semantic
segmentation, here a Fully-Convolutional-Neural-Network (FCN)~\cite{yu2015multi}, is part of our new edge-detection approach. Again, advances in semantic segmentation may improve the performance of this component, as well. \\

{\bf Our Contributions} in short form are:\\
\indent $\bullet$ We propose a novel paradigm for instance-aware semantic segmentation, which has different pros and cons than existing approaches. In our approach, we only train classifiers for semantic segmentation and instance-edge detection, and not directly any classifier for dealing with global properties of an instance, such as shape.  \\
\indent $\bullet$ We propose a novel MultiCut formulation that reasons globally about the optimal partitioning of an image into instances. \\
\indent $\bullet$ We propose a new FCN-based architecture for instance-aware edge detection. \\
\indent $\bullet$ We validate experimentally that our approach achieves the best result, among all published methods, and performs particularly well for rare object classes.\\

\section{Related Work}\label{sec:related_works}

\myparagraph{Proposal-based methods.} This group of methods uses detection or a proposal generation mechanism as a subroutine in the instance-aware segmentation pipeline.

Several recent methods decompose the instance-aware segmentation problem into a detection stage and a foreground/background segmentation stage~\cite{dai2015instance,hariharan2015hypercolumns}. These methods propose an end-to-end training that incorporates all parts of the model. In addition, non-maximal suppression (NMS) may be employed as a post-processing step. A very similar approach generates proposals using e.g. MCG~\cite{arbelaez2014multiscale} and then, in the second stage, a different network classifies these proposals~\cite{Cordts2016Cityscapes,hariharan2014simultaneous,dai2015convolutional,chen2015multi}. 

Several methods produce proposals for instance segmentations and combine them, based on learned scores~\cite{liang2015reversible,DeepMask,SharpMask} or generate parts of instances and then combine them~\cite{dai2016instance,liumulti}.

Although the  proposal-based methods show state-of-the-art performance on important challenges,  Pascal VOC2012~\cite{everingham2010pascal} and MSCOCO~\cite{lin2014microsoft},  they are limited by the quality of the used detector or proposal generator. Our method is, in turn, dependent on the quality of the used semantic segmentation. However, for the latter a considerable amount of research exists with high quality results.

\myparagraph{Proposal-free methods.} 
Recently, a number of alternative techniques to proposal-based approaches have been suggested in the literature. These methods explore different decompositions of instance-aware semantic segmentation followed by a post-processing step that assembles results.

In~\cite{uhrig2016pixel} the authors propose a template matching scheme for instance-aware segmentation based on three modalities: predicted semantic segmentation, depth estimation, and per-pixel direction estimation with respect to the center of the corresponding instance. The approach requires depth data for training and does not perform well on highly occluded objects.

Another work, which focuses on instance segmentation of cars~\cite{zhang2015monocular,zhang2015instance} employs a conditional random field that reasons about instances using multiple overlapping outputs of an FCN. The latter predicts a fixed number of instances and their order within the receptive field of the FCN, i.e.~for each pixel, the FCN predicts an ID of the corresponding instance or background label. 
However, in these methods the maximal number of instances per image must be fixed in advance. A very large number may have a negative influence on the system performances. Therefore, this method may not be well-suited for the CityScapes dataset, where the number of instances varies considerably among images.

In~\cite{wu2016bridging} the authors predict the bounding box of an instance for each pixel, based on instance-agnostic semantic segmentation. A post-processing step filters out the resulting instances.  

Recurrent approaches produce instances one-by-one. In~\cite{ren2016end} an attention-based recurrent neural network is presented. In~\cite{romera2015recurrent} an LSTM-based~\cite{hochreiter1997long} approach is proposed.

The work~\cite{liang2015proposal} presents a proposal-free network that produces an instance-agnostic semantic segmentation, number of instances for the image, and a per-pixel bounding box of the corresponding instance. The resulting instance segmentation is obtained by clustering. The method is highly sensitive to the right prediction of the number of instances.

We also present a proposal-free method. However, ours is very different in paradigm. To infer instances, it combines semantic segmentation and object boundary detection via global reasoning.

\section{InstanceCut}\label{sec:method}

\subsection{Overview of the proposed framework}

\begin{figure}[ht!]
	\centering
	\includegraphics[width=0.84\linewidth]{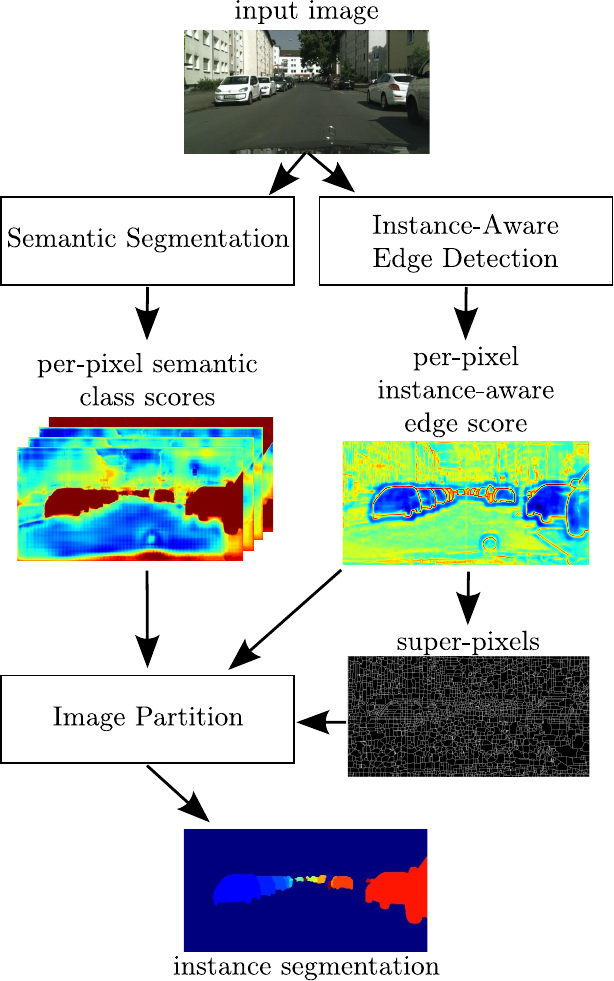}
	\caption{{\bf Our InstanceCut pipeline - Overview.} Given an input image, two independent branches produce the per-pixel semantic class scores and per-pixel instance-aware edge scores. The edge scores are used to extract superpixels. The final image partitioning block merges the superpixels into connected components with a class label assigned to each component. The resulting components correspond to object instances and background.}
	\label{fig:overview}
\end{figure}

We begin with presenting a general pipeline of our new InstanceCut framework (see Fig.~\ref{fig:overview}) and then describe each component in detail. The first two blocks of the pipeline are processed independently: {\em semantic segmentation} and {\em instance-aware edge detection} operate directly on the input image. The third, {\em image partitioning block}, reasons about instance segmentation on the basis of the output provided by the two blocks above.

More formally, the semantic segmentation block (Section~\ref{sec:semantic}) outputs a log-probability of a semantic class $a_{i, l}$ for each class label $l \in \SL = \{0, 1 \ldots, L\}$ and each pixel $i$ of the input image. We call $a_{i, l}$, {\em per-pixel semantic class scores}.  Labels $1, \ldots, L$ correspond to different semantic classes and $0$ stands for background. 

Independently, the instance-aware edge detection (Section~\ref{sec:edge}) outputs log-probabilities $b_{i}$ of an object boundary for each pixel $i$.
In other words, $b_{i}$ indicates how likely it is that pixel $i$ touches an object boundary. We term $b_{i}$ as {\em a per-pixel instance-aware edge score}. Note that these scores are class-agnostic.

Finally, the image partitioning block outputs the resulting instance segmentation, obtained using the semantic class scores and the instance-aware edge scores. We refer to Section~\ref{sec:ilp} for a description of the corresponding optimization problem. To speed-up optimization, we reduce the problem size by resorting to a superpixel image. 
For the superpixel extraction we utilize the well-known watershed technique~\cite{vincent1991watersheds}, which is run directly on the edge scores. This approach efficiently ensures that the extracted superpixel boundaries are aligned with boundaries of the instance-aware edge scores. 

\subsection{Semantic Segmentation}\label{sec:semantic}

Recently proposed semantic segmentation frameworks are mainly based on the fully convolution network (FCN) architecture. Since the work~\cite{long2015fully}, many new FCN architectures were proposed for this task~\cite{yu2015multi,ghiasi2016laplacian}. Some of the methods utilize a conditional random field (CRF) model on top of an FCN~\cite{chen2016deeplab,lin2015efficient}, or incorporate CRF-based mechanisms directly into a network architecture~\cite{liu2015semantic,zheng2015conditional,schwing2015fully}. Current state-of-the-art methods report around $78\%$ mean Intersection-over-Union (IoU) for the CityScapes dataset~\cite{Cordts2016Cityscapes} and about $82\%$ for the PASCAL VOC2012 challenge~\cite{everingham2010pascal}. 
Due to the recent progress in this field, one may say that with a sufficiently large dataset, with associated dense ground truth annotation, an FCN is able to predict semantic class for each pixel with high accuracy.

In our experiments, we employ two publicly available pre-trained FCNs: Dilation10~\cite{yu2015multi} and LRR-4x~\cite{ghiasi2016laplacian}. These networks have been trained by the respective authors and we can also use them as provided, without any fine-tuning. Note, that we also use the CNN-CRF frameworks~\cite{zheng2015conditional,chen2016deeplab} with dense CRF~\cite{koltun2011efficient}, since dense CRF's output can also be treated as the log-probability scores $a_{i, l}$.

Since our image partitioning framework works on the superpixel level we transform the pixel-wise semantic class scores $a_{i, l}$ to the superpixel-wise ones $a_{u, l}$ (here $u$ indexes the superpixels) by averaging the corresponding pixels' scores.

\subsection{Instance-Aware Edge Detection}\label{sec:edge}

Let us first review existing work, before we describe our approach. Edge detection (also know as {\em boundary detection}) is a very well studied problem in computer vision. The classical results were obtained already back in the $80$'s~\cite{canny1986computational}. More recent methods are based on spectral clustering~\cite{shi2000normalized,arbelaez2011contour,arbelaez2014multiscale,isola2014crisp}. These methods perform global inference on the whole image. An alternative approach suggests to treat the problem as a per-pixel classification task~\cite{lim2013sketch,dollar2015fast}. Recent advances in deep learning have made this class of methods especially efficient, since they automatically obtain rich feature representation for classification~\cite{ganin2014n,kivinen2014visual,shen2015deepcontour,
bertasius2015deepedge,bertasius2015high,xie2015holistically,bertasius2015semantic}.

The recent per-pixel classification method~\cite{bertasius2015semantic} constructs features, which are based on an FCN trained for semantic segmentation on Pascal VOC 2012~\cite{everingham2010pascal}. The method produces state-of-the-art edge detection performance on the BSD500 dataset~\cite{amfm_pami2011}. 
The features for each pixel are designed as a concatenation of intermediate FCN features, corresponding to that particular pixel. 
The logistic regression trained on these features, followed by non-maximal suppression, outputs a per-pixel edge probability map. The paper suggests that the intermediate features of an FCN trained for semantic segmentation form a strong signal for solving the edge detection problem. Similarly constructed features also have been used successfully for other dense labelling problems~\cite{hariharan2015hypercolumns}.

\begin{figure*}
	\centering
	\includegraphics[width=0.98\linewidth]{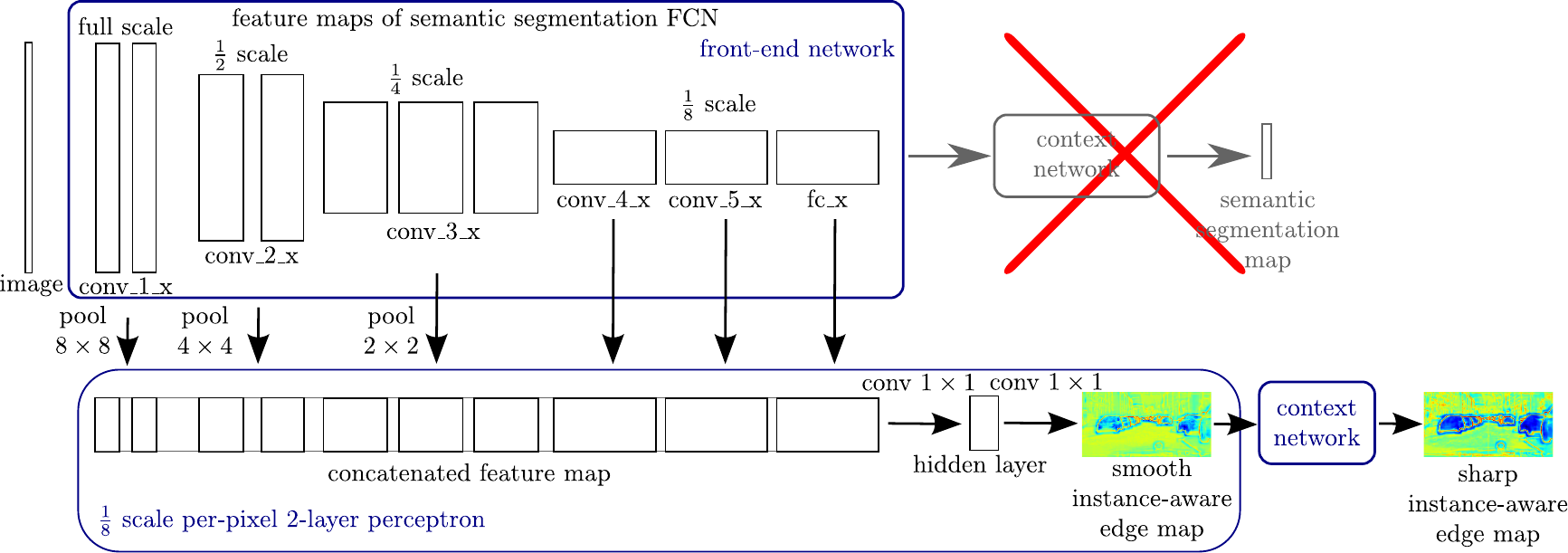}
	\caption{{\bf Instance-aware edge detection block.} The semantic segmentation FCN is the front-end part of the network~\cite{yu2015multi} trained for semantic segmentation on the same dataset. Its intermediate feature maps are downsampled, according to the size of the smallest feature map, by a max-pooling operation with an appropriate stride. The concatenation of the downsampled maps is used as a feature representation for a per-pixel 2-layer perceptron. The output of the perceptron is refined by a context network of Dilation10~\cite{yu2015multi} architecture.}
\label{fig:detector}
\end{figure*}

For datasets like BSDS500~\cite{amfm_pami2011} most works consider general edge detection problem, where annotated edges are class- and instance-agnostic contours. 
In our work the instance-aware edge detection outputs a probability for each pixel, whether it touches a boundary.
This problem is more challenging than canonical edge detection, since it requires to reason about contours and semantics jointly, distinguishing the true objects' boundaries and other not relevant edges, e.g. inside the object or in the background.
Below (see Fig.~\ref{fig:detector}), we describe a new network architecture for this task that utilizes the idea of the intermediate FCN features concatenation. 

As a base for our network we use an FCN that is trained for semantic segmentation on the dataset that we want to use for object boundary prediction. In our experiments we use a pre-trained Dilation10~\cite{yu2015multi} model, however, our approach is not limited to this architecture and can utilize any other FCN-like architectures. We form a per-pixel feature representation by concatenating the intermediate feature maps of the semantic segmentation network. This is based on the following intuition: during inference, the semantic segmentation network is able to identify positions of transitions between semantic classes in the image. Therefore, its intermediate features are likely to contain a signal that helps to find the borders between classes. We believe that the same features can be useful to determine boundaries between objects. 

Commonly used approaches~\cite{bertasius2015semantic,hariharan2015hypercolumns} suggest upscaling feature maps that have a size which is smaller than the original image to get per-pixel representation. However, in our experiments such an approach produces thick and over-smooth edge scores. This behavior can be explained by the fact that the most informative feature maps have an $8$ times smaller scale than the original image. Hence, instead of upscaling, we downscale all feature maps to the size of the smallest map. Since the network was trained with rectified linear unit (ReLU) activations, the active neurons tends to output large values, therefore, we use max-pooling with a proper stride for the downscaling, see Fig.~\ref{fig:detector}.

The procedure outputs the downscaled feature maps (of a {\em semantic segmentation FCN}, see Fig.~\ref{fig:detector}) that are {\em concatenated} to get the downscaled per-pixel {\em feature map}. We utilize a {\em 2-layer perceptron} that takes this feature map as input and outputs log-probabilities for edges ({\em smooth instance-aware edge map}, see Fig.~\ref{fig:detector}). The perceptron method is the same for all spatial positions, therefore, it can be represented as two layers of $1\times1$ convolutions with the ReLU activation in between. 

In our experiments we have noticed that the FCN gives smooth edge scores. Therefore, we apply {\em a context network}~\cite{yu2015multi} that refines the scores making them sharper. The new architecture is an FCN, i.e. it can be applied to images of arbitrary size, it is differentiable and has a single loss at the end. Hence, straightforward end-to-end training can be applied for the new architecture. We upscale the resulting output map to match an input image size. 

Since the image partition framework, that comes next, operates on super-pixels, we need to transform the per-pixel edge scores $b_{i}$ to edge scores $b_{u, v}$ for each pair $\{u,v\}$ of neighboring superpixels. We do this by averaging all scores of of those pixels that touch the border between $u$ and $v$.

In the following, we describe an efficient implementation of the 2-layer perceptron and also discuss our training data for the boundary detection problem.

\myparagraph{Efficient implementation.} In our experiments, the input for the 2-layer perceptron contains about 13k features per pixel. Therefore, the first layer of the perceptron consumes a lot of memory. It is, however, possible to avoid this by using a more efficient implementation. Indeed, the first layer of the perceptron is equivalent to the summation of outputs of multiple $1\times1$ convolutions, which are applied to each feature map independently. For example, {\tt conv\_1} is applied to the feature maps from the {\tt conv\_1\_x } intermediate layer, {\tt conv\_2} is applied to the feature maps from {\tt conv\_2\_x} and its output is summed up with the output of {\tt conv\_1}, etc. This approach allows reducing the memory consumption, since the convolutions can be applied during evaluation of the front-end network.

\myparagraph{Training data.} 
Although it is common for ground truth data that object boundaries lie in-between pixels, we will use in the following the notion that a boundary lies on a  pixel. Namely,
we will assume that a pixel $i$ is labeled as a boundary if there is a neighboring pixel $j$, which is assigned to a different object (or background). Given the size of modern images, this boundary extrapolation does not affect performance. As a ground truth for boundary detection we use the boundaries of object instances presented in CityScapes~\cite{Cordts2016Cityscapes}.

As mentioned in several previous works~\cite{xie2015holistically,bertasius2015high}, highly unbalanced ground truth (GT) data heavily harms the learning progress. For example, in BSDS500~\cite{amfm_pami2011} less than $10\%$ of pixels on average are labeled as edges. Our ground truth data is even more unbalanced: since we consider the object boundaries only, less than $1\%$ of pixels are labeled as being an edge. We employ two techniques to overcome this problem of training with unbalanced data: {\em a balanced loss function}~\cite{xie2015holistically,hwang2015pixel} and {\em pruning of the ground truth data}.

The balanced loss function~\cite{xie2015holistically,hwang2015pixel} adds a coefficient to the standard log-likelihood loss that decreases the influence of errors with respect to classes that have a lot of training data. That is, for each pixel $i$ the balanced loss is defined as
\begin{align}
	loss(p_{edge}, y^{GT}) = & \llbracket y^{GT} = 1 \rrbracket \log(p_{edge})\\ \nonumber
	& + \alpha \llbracket y^{GT} = 0 \rrbracket \log(1 - p_{edge}) \,,
\end{align}
where $p_{edge}=1/(1-e^{-b_i})$ is the probability of the pixel~$i$ to be labeled as an edge, $y^{GT}$ is the ground truth label for $i$ (the label $1$ corresponds to an edge), and $\alpha = N_{1} / N_{0}$ is the balancing coefficient. 
Here, $N_{1}$ and $N_{0}$ are numbers of pixels labeled, respectively, as $1$ and $0$ in the ground truth.

\begin{figure}
	\centering
	\includegraphics[width=0.45\linewidth]{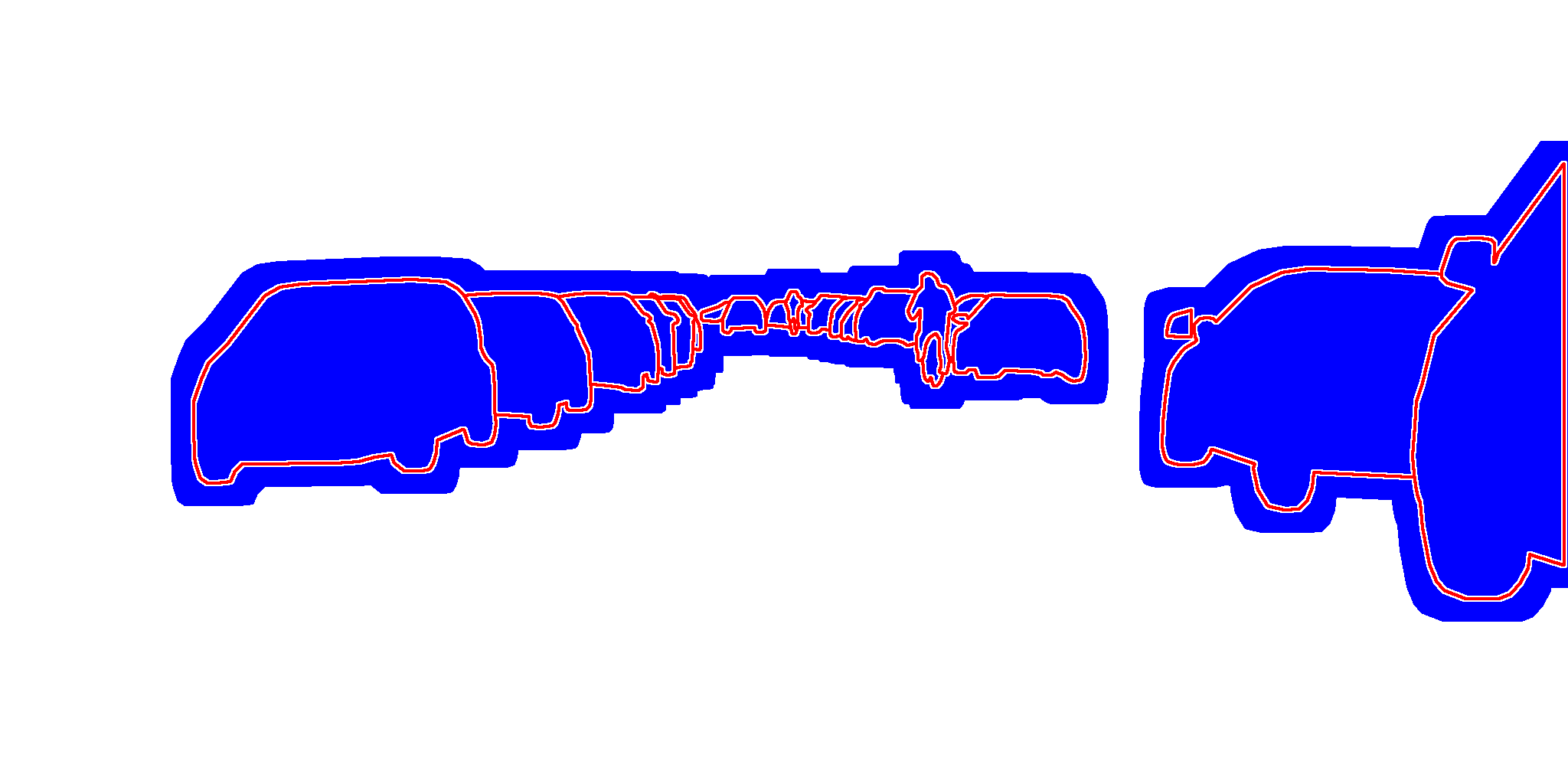}
	\hspace{0.05cm}
	\includegraphics[width=0.45\linewidth]{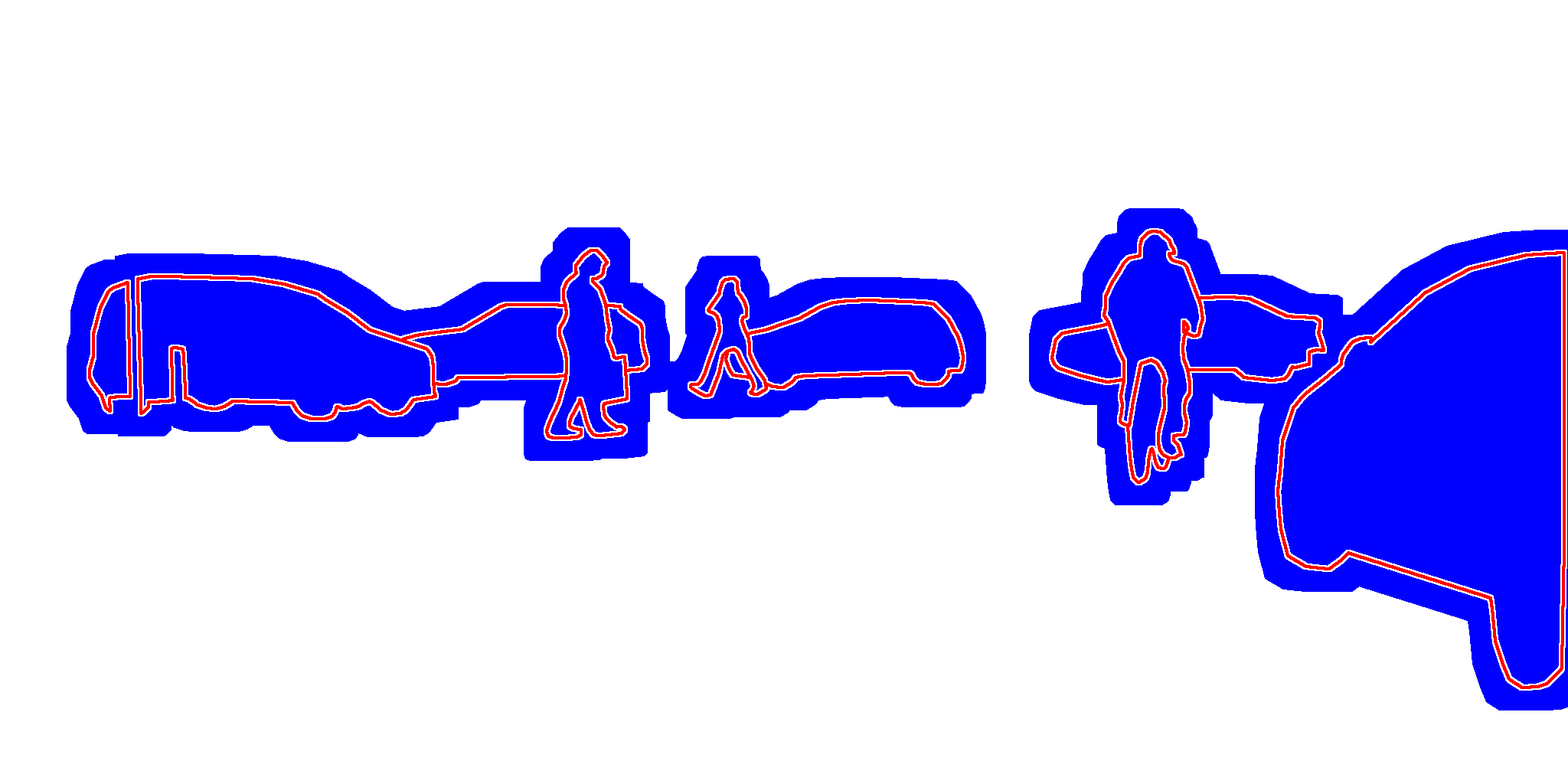}
	\caption{Ground truth examples for our instance-aware edge detector. Red indicates pixels that are labeled as edges, blue indicates background, i.e. no edge and white pixels are ignore.}
	\label{fig:gt_edge_data}
\end{figure}

Another way to decrease the effect of unbalanced GT data is to subsample the GT pixels, see e.g.~\cite{bertasius2015semantic}. Since we are interested in instance-aware edge detection and combine its output with our semantic segmentation framework, a wrong edge detection, which is far from the target objects (for example, in the sky) does not harm the overall performance of the InstanceCut framework. Hence, we consider a pixel to be labeled as background for the instance-aware edge detection if and only if it lies inside the target objects, or in an area close to it, see Fig.~\ref{fig:gt_edge_data} for a few examples of the ground truth data for the CityScapes dataset~\cite{Cordts2016Cityscapes}. In our experiments, only $6.8\%$ of the pixels are labeled as object boundaries in the pruned ground truth data.

\subsection{Image Partition}\label{sec:ilp}

Let $V$ be the set of superpixels extracted from the output of the instance-aware edge detection block and $E\subseteq {V\choose 2}$ be the set of neighboring superpixels, i.e., those having a common border.

With the methods described in Sections~\ref{sec:semantic} and~\ref{sec:edge} we obtain: \\
\indent $\bullet$ Log-probabilities $\alpha_{u,l}$ of all semantic labels $l\in \SL$ (including background) for each superpixel $u\in V$. \\
\indent $\bullet$ Log-probabilities $b_{u,v}$ for all pairs of neighbouring superpixels $\{u,v\}\in E$, for having a cutting edge.  \\
\indent $\bullet$ Prior log-probabilities of having a boundary between any two (also equal) semantic classes $\beta_{l,l'}$, for any two labels $l,l'\in \SL$. In particular, the weight $\beta_{l,l}$ defines, how probable it is that two neighboring super-pixel have the same label $l$ and belong to different instances. We set $\beta_{0,0}$ to $-\infty$, assuming there are no boundaries between superpixels labeled both as background.

We want to assign a single label to each superpixel and have close-contour boundaries, such that if two neighboring superpixels belong to different classes, there is always a boundary between them.

Our problem formulation consists of two components: (i) a conditional random field model~\cite{kappes-2015-ijcv} and (ii) a graph partition problem, known as MultiCut~\cite{Chopra1993} or correlation clustering~\cite{Bansal2004}. In a certain sense, these two problems are coupled together in our formulation. Therefore, we first briefly describe each of them separately and afterwards consider their joint formulation.

\myparagraph{Conditional Random Field (CRF).} Let us, for now, assume that all $\beta_{l,l}=-\infty$, $l\in \SL$, i.e., there can be no boundary between superpixels assigned the same label. In this case our problem is reduced to the following famous format:
Let $G=(V,E)$ be an undirected graph. A finite set of labels $\SL$ is associated with each node. With each label $l$ in each node $v$ a vector $\alpha_{v,l}$ is associated, which denotes {\em the score} of the label assigned to the node. Each pair of labels $l,l'$ in neighbouring nodes $\{u,v\}$ is assigned a score 
$c_{u,v,l,l'}:=
\begin{cases}
 b_{u,v} + \beta_{l,l'}, & l\neq l'\\
 0, & l = l'
\end{cases}
$. \\ 
The vector $\mathbf{l}\in \SL^{|V|}$ with coordinates $l_u$, $u\in V$ being labels assigned to each node is called {\em a labeling}.
{\em The maximum a posteriori inference} problem for the CRF is defined above reads
\begin{equation}\label{equ:map-inference}
 \max_{\mathbf{l}\in L^{|V|}}\sum_{u\in V}\alpha_{u,l_u}+\sum_{uv\in E}c_{u,v,l_u,l_v}\,.
\end{equation}
A solution to this problem is a usual (non-instance-aware) semantic segmentation, if we associate the graph nodes with superpixels and the graph edges will define their neighborhood. 

For the MultiCut formulation below, we will require a different representation of the problem~\eqref{equ:map-inference}, in a form of {\em an integer quadratic problem}. 
Consider binary variables $x_{u,l}\in\{0,1\}$ for each node $u\in V$ and label $l\in \SL$. The equality $x_{u,l}=1$ means that label $l$ is assigned to the node $u$. The problem~\eqref{equ:map-inference} now can be rewritten as follows:
\begin{align}
 & \max_{\mathbf{x}}\sum_{u\in V}\sum_{l\in \SL}\alpha_{u,l}x_{u,l}+\sum_{uv\in E}\sum_{l\in \SL}\sum_{l'\in \SL}c_{u,v,l,l'}x_{u,l}x_{v,l'}\nonumber\\
 & \text{s.t.}\ \begin{cases} \label{equ:iqp-map-inference}
                 x_{u,l}\in\{0,1\}, & u\in V, l\in \SL\\
                 \sum_{l\in L}x_{u,l}=1, & u\in V\,.
                \end{cases}
\end{align}
The last constraint in~\eqref{equ:iqp-map-inference} is added to guarantee that each node is assigned exactly one label. Although the problem~\eqref{equ:iqp-map-inference} is NP-hard in general, it can be efficiently (and often exactly) solved for many practical instances appearing in computer vision, see~\cite{kappes-2015-ijcv} for an overview.

\myparagraph{MultiCut Problem.} 
Let us now assume a different situation, where all nodes have already got an assigned semantic label and all that we want is to {\em partition} each connected component (labeled with a single class) 
into connected regions corresponding to instances. Let us assume, for instance, that all superpixels of the component have a label $l$. This task has an elegant formulation as {\em a MultiCut} problem~\cite{Chopra1993}:

Let $G=(V,E)$ be an undirected graph, with the scores $\theta_{u,v}:=b_{u,v}+\beta_{l,l}$ assigned to the graph edges. Let also $\dot{\cup}$ stand for a disjoint union of sets.
{\em The MultiCut} problem (also known as correlation clustering)  is to find a partitioning $(\Pi_1,\ldots,\Pi_k)$, $\Pi_i\subseteq V$, $V=\dot{\cup}_{i=1}^{k}\Pi_i$ of the graph vertices, such that the total score of edges connecting different components is maximized. The number $k$ of components is not fixed but is determined by the algorithm itself. Although the problem is NP-hard in general, there are efficient approximate solvers for it, see e.g.~\cite{beier2014cut,kernighan1970efficient,keuper2015efficient}.

In the following, we will require a different representation of the MultiCut problem, in form of {\em an integer linear problem}. 
To this end, we introduce a binary variable $y_e= y_{u,v}\in\{0,1\}$ for each edge $e=\{u,v\}\in E$. This variable takes the value $1$, if $u$ and $v$ belong to different components, i.e.\ $u\in\Pi_i$, $v\in\Pi_j$ for some $i\ne j$. Edges $\{u,v\}$ with $y_{u,v}=1$ are called {\em cut} edges. The vector $\mathbf{y}\in\{0,1\}^{|E|}$ with coordinates $y_e$, $e\in E$ is called {\em a MultiCut}. Let~$C$ be the set of all cycles of the graph $G$. It is a known result from combinatorial optimization~\cite{Chopra1993} that the MultiCut problem can be written in the following form:
\begin{align}\label{equ:multicut-def}
 \hspace{-10pt}\max_{\mathbf{y}\in\{0,1\}^{|E|}}\hspace{-3pt}\sum_{\{u,v\}\in E}\hspace{-5pt}\theta_{u,v}y_{u,v}\,,\ \ \text{s.t.}\
 \forall C\ \forall e'\in C \colon \hspace{-5pt}\sum_{e\in C\backslash\{e'\}}\hspace{-5pt}y_{e} \ge y_{e'}\,.  
\end{align}
Here, the objective directly maximizes the total score of the edges and the inequality constraints basically force each cycle to have none or at least two cut edges. These {\em cycle constraints} ensure that the set of cut edges actually defines a partitioning. Obviously, the cut edges correspond to boundaries in our application.

\myparagraph{Our InstanceCut Problem.} Let us combine both subproblems: We want to {\em jointly} infer both the semantic labels {\em and} the partitioning of each semantic segment, with each partition component defining an object instance. To this end, consider our InstanceCut problem~\eqref{equ:InstanceCut-objective}-\eqref{equ:InstanceCut-coupling} below:
\begin{align}\label{equ:InstanceCut-objective}
 &\hspace{-8pt}\max_{\mathbf{x}\in \{0,1\}^{|V||\SL|}\atop \mathbf{y}\in\{0,1\}^{|E|}}
 \sum_{u\in V}\sum_{l\in \SL}\alpha_{u,l}x_{u,l}\\
 &\hspace{50pt} +w \sum_{uv\in E}\sum_{l\in \SL}\sum_{l'\in \SL}(b_{u,v}+\beta_{l,l'})x_{u,l}x_{v,l'}y_{u,v}\nonumber \\
 & \sum_{l\in \SL}x_{u,l}=1,\ u\in V \label{equ:InstanceCut-uniqueness}\\
 & \forall e'\in C \colon \hspace{-5pt}\sum_{e\in C\backslash\{e'\}}\hspace{-5pt}y_{e} \ge y_{e'} \label{equ:InstanceCut-cycle}\\
 &\left. \label{equ:InstanceCut-coupling}
 \begin{array}{l}
 x_{u,l} - x_{v,l} \le y_{uv} \\
 x_{u,l} - x_{v,l} \le y_{uv} 
 \end{array}
 \right\},\ \{u,v\}\in E,\ l\in \SL\,. 
\end{align}
Objective~\eqref{equ:InstanceCut-objective} and inequalities~\eqref{equ:InstanceCut-uniqueness}-\eqref{equ:InstanceCut-cycle} are obtained directly from merging problems~\eqref{equ:iqp-map-inference} and~\eqref{equ:multicut-def}. We also introduced the parameter $w$ that balances the modalities. Additional constraints~\eqref{equ:InstanceCut-coupling} are required to guarantee that as soon as two neighboring nodes $u$ and $v$ are assigned different labels, the corresponding edge $y_{u,v}$ is cut and defines a part of an instance boundary. Two nodes $u$ and $u$ are assigned different labels if at most one of the variables $x_{u,l}$, $x_{v,l}$ takes value $1$. In this case, the largest left-hand side of one of the inequalities~\eqref{equ:InstanceCut-coupling} is equal to $1$ and therefore $y_{u,v}$ must be cut. The problem related to~\eqref{equ:InstanceCut-objective}-\eqref{equ:InstanceCut-coupling} was considered in~\cite{hamprecht2014asymmetric} for foreground/background segmentation.

Although the problem~\eqref{equ:InstanceCut-objective}-\eqref{equ:InstanceCut-coupling} is NP-hard and it contains a lot of hard constraints, there exists an efficient approximate solver for it~\cite{solver}, which we used in our experiments. For solving the problem over 3000 nodes (superpixels) and $9$ labels (segment classes) it required less than a second on average.

\section{Experiments}\label{sec:experiments}

\begin{figure}
	\centering
	\includegraphics[width=0.85\linewidth]{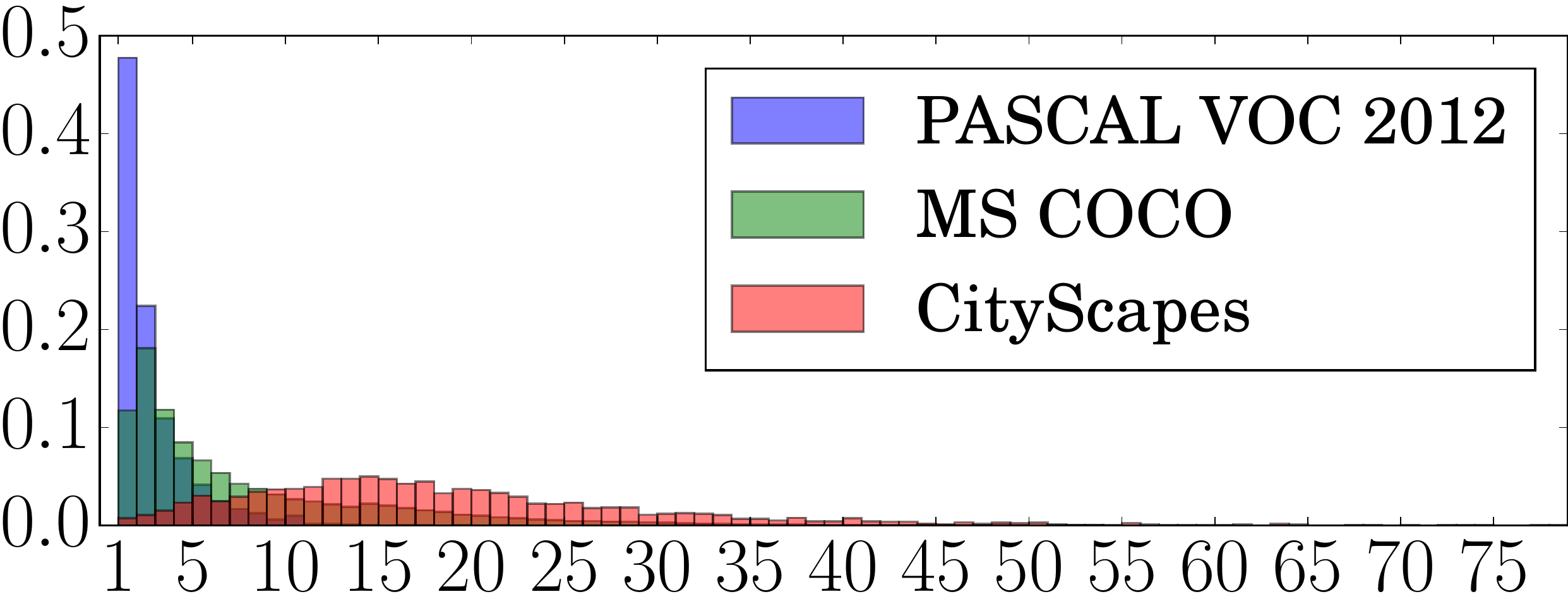}
	\caption{The histograms shows distribution of number of instances per image for different datasets. For illustrative reasons we cut long tails of CityScapes and MS COCO. We use CityScapes dataset since it contains significantly more instances per image.}
	\label{fig:hist}
\end{figure}

\myparagraph{Dataset.} There are three main datasets with full annotation for the instance-aware semantic segmentation problem: PASCAL VOC2012~\cite{everingham2010pascal},  MS COCO~\cite{lin2014microsoft} and CityScapes~\cite{Cordts2016Cityscapes}. We select the last one for our experimental evaluation for several reasons: (i) CityScapes has very fine annotation with precise boundaries for the annotated objects, whereas MS COCO has only coarse annotations, for some objects, that do not coincide with the true boundaries. Since our method uses an edge detector, it is important to to have precise object boundaries for training. (ii) The median number of instances per image in CityScapes is $16$, whereas PASCAL VOC has $2$ and MS COCO has $4$. For this work a larger number is more interesting. The distribution of the number of instances per image for different datasets is shown in Fig.~\ref{fig:hist}. (iii) Unlike other datasets, CityScapes' annotation is dense, i.e.~all foreground objects are labeled.

The CityScape dataset has $5000$ street-scene images recorded by car-mounted cameras: $2975$ images for training, $500$ for validation and $1525$ for testing. There are $8$ classes of objects that have an instance-level annotation in the dataset: person, rider, car, truck, bus, train, motorcycle, bicycle. All images have the size of $1024 \times 2048$ pixels.

\begin{figure*}	
	\begin{subfigure}{0.33\linewidth}
		\includegraphics[width=0.99\linewidth]{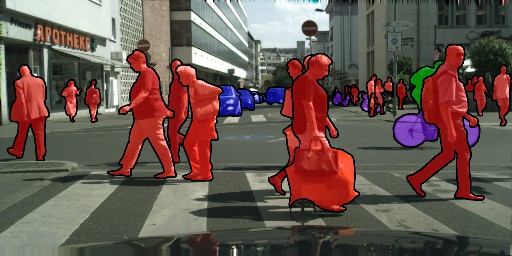}
	\end{subfigure}%
	\hspace{0.03cm}
	\begin{subfigure}{0.33\linewidth}
		\includegraphics[width=0.99\linewidth]{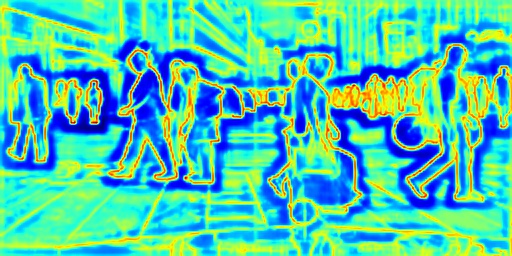}
	\end{subfigure}%
	\hspace{0.01cm}
	\begin{subfigure}{0.33\linewidth}
		\includegraphics[width=0.99\linewidth]{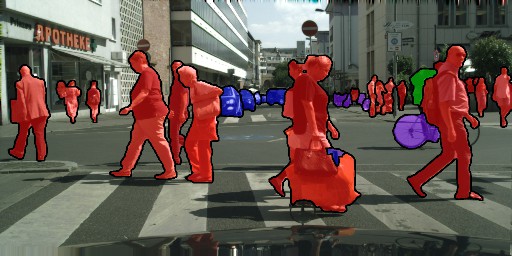}
	\end{subfigure}%
	\\[0.03cm]
	\begin{subfigure}{0.33\linewidth}
		\includegraphics[width=0.99\linewidth]{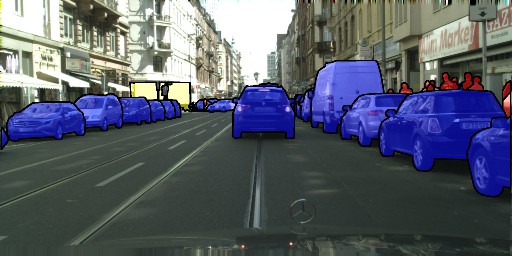}
	\end{subfigure}%
	\hspace{0.01cm}
	\begin{subfigure}{0.33\linewidth}
		\includegraphics[width=0.99\linewidth]{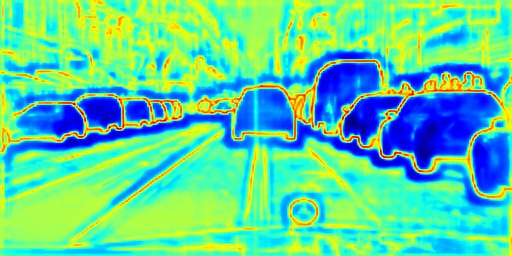}
	\end{subfigure}%
	\hspace{0.01cm}
	\begin{subfigure}{0.33\linewidth}
		\includegraphics[width=0.99\linewidth]{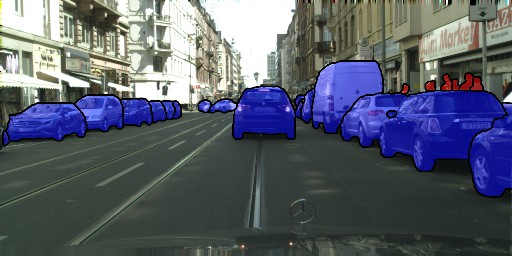}
	\end{subfigure}%
	\\[0.03cm]
	\begin{subfigure}{0.33\linewidth}
		\includegraphics[width=0.99\linewidth]{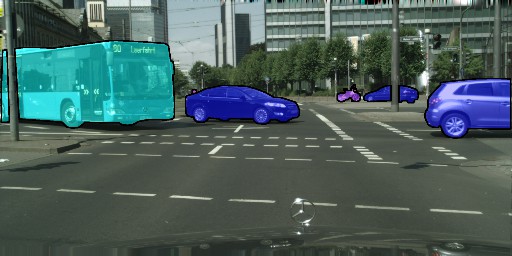}
		\caption{Ground truth}
	\end{subfigure}%
	\hspace{0.03cm}
	\begin{subfigure}{0.33\linewidth}
		\includegraphics[width=0.99\linewidth]{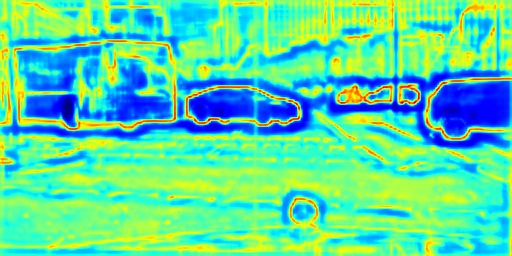}
		\caption{Edges map}
	\end{subfigure}%
	\hspace{0.03cm}
	\begin{subfigure}{0.33\linewidth}
		\includegraphics[width=0.99\linewidth]{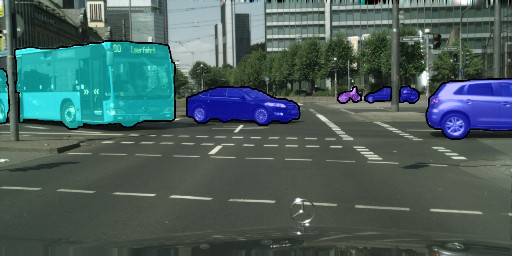}
		\caption{InstanceCut prediction}
	\end{subfigure}%
	\vspace{-0.03cm}
	\caption{Qualitative results of InstanceCut framework. Left column contains input images with the highlighted ground truth instances. Middle column depicts per-pixel instance-aware edge log-probabilities and the last column shows the results of our approach. Note that in the last example the bus and a car in the middle are separated by a lamp-post, therefore, our method returns two instances for the objects.}
	\label{fig:results}
\end{figure*}

\myparagraph{Training details.}
For the semantic segmentation block in our framework we test two different networks, which have publicly available trained models for CityScapes: Dilation10~\cite{yu2015multi} and LRR-4x~\cite{ghiasi2016laplacian}. The latter is trained using the additional coarsely annotated data, available in CityScapes. Importantly, CityScapes has $19$ different semantic segmentation classes (and only $8$ out of them are considered for instance segmentation) and both networks were trained to segment all these classes. We do not retrain the networks and directly use the log-probabilities for the $8$ semantic classes, which we require. For the background label we take the maximum over the log-probabilities of the remaining semantic classes.

As an initial semantic segmentation network for the instance-aware edge detection block we use Dilation10~\cite{yu2015multi} pre-trained on the CityScapes. We exactly follow the training procedure described in the original paper~\cite{yu2015multi}. That is, we pre-train first the front-end module with the 2-layer perceptron on top. Then we pre-train the context module of the network separately and, finally, train the whole system end-to end. All the stages are trained with the same parameters as in~\cite{yu2015multi}. In our experiments the 2-layer perceptron has $16$ hidden neurons. On the validation set the trained detector achieves $97.2\%$ AUC ROC.

Parameters $w$ (see~\eqref{equ:InstanceCut-objective}) and $\beta_{l, l'}$, for all $l, l' \in \SL$, in our InstanceCut formulation \eqref{equ:InstanceCut-objective} are selected via 2-fold cross-validation. 
Instead of considering different $\beta_{l, l'}$ for all pairs of labels, we group them into two classes: 'big' and 'small'. All $\beta_{l, l'}$, where either $l$ or $l'$ corresponds to a (physically) big object, i.e., train, bus, or truck, are set to $\beta_{big}$. All other $\beta_{l, l'}$ are set to $\beta_{small}$.
Therefore, our parameter space is only $3$ dimensional and is determined by the parameters $w$, $\beta_{small}$ and $\beta_{big}$. 

\begin{table}[t]
\small
\tabcolsep=0.20cm
\centering
\begin{tabular}{l|cccc}
\toprule
                      & AP            & AP$50\%$      & AP$100$m      & AP$50$m       \\ \midrule
MCG+R-CNN~\cite{Cordts2016Cityscapes}    & 4.6           & 12.9          & 7.7           & 10.3          \\
Uhrig et al.~\cite{uhrig2016pixel} & 8.9           & 21.1          & 15.3          & 16.7          \\
InstanceCut           & \textbf{13.0} & \textbf{27.9} & \textbf{22.1} & \textbf{26.1} \\ \bottomrule
\end{tabular}
\vspace{-0.03cm}
\caption{CityScapes results for instance-aware semantic segmentation on the test set. The table contains only published results.\protect\footnotemark }
\label{tab:results_class}
\end{table}

\myparagraph{Instance-level results - quantitative and qualitative.}
We evaluated our method using $4$ metrics that are suggested by the CityScapes benchmark: AP, AP$50$\%, AP$100$m and AP$50$m. We refer to the webpage of the benchmark for a detailed description.

The InstanceCut framework with Dilation10~\cite{yu2015multi} as the semantic segmentation block gives AP~$=14.8$ and AP$50\%=30.7$ on the validation part of the dataset. When we replace Dilation10 by LRR-4x~\cite{ghiasi2016laplacian} for this block the performance improves to AP~$=15.8$ and AP$50\%=32.4$, on the validation set.

Quantitative results for the test set are provided in Table~\ref{tab:results_class}. We compare our approach to all published methods that have results for this dataset. Among them our method shows the best performance, despite its simplicity.

\footnotetext{In the days, and hours, before the submission deadline other methods appeared in the table, which remain unpublished so far. For this reason we exclude them from the comparison.} We refer to the supplementary material for the class-level results. A few qualitative results are shown in Fig.~\ref{fig:results}.

\section{Conclusion}\label{sec:conclusion}

We have proposed an alternative paradigm for instance-aware semantic segmentation. The paradigm represents the instance segmentation problem by a combination of two modalities: instance-agnostic semantic segmentation and instance-aware boundaries. We have presented a new framework that utilize this paradigm. The modalities are produced by FCN networks. The standard FCN model is used for semantic segmentation, whereas a new architecture is proposed for object boundaries. The modalities are combined are combined by a novel MultiCut framework, which reasons globally about instances. The proposed framework achieves the best results amongst all published methods for the challenging CityScapes dataset\cite{Cordts2016Cityscapes}. 

\section{Acknowledgments}

We would like to thank Michael Figurnov for helpful discussions. This project has received funding from the European Research Council (ERC) under the European Union’s
Horizon 2020 research and innovation programme (grant agreement No 647769).

{\small
\bibliographystyle{ieee}
\bibliography{egbib}
}

\clearpage
\appendix

\begin{center}
\textbf{\large Supplementary materials}
\end{center}

\begin{figure*}	
	\begin{subfigure}{0.33\linewidth}
		\includegraphics[width=0.99\linewidth]{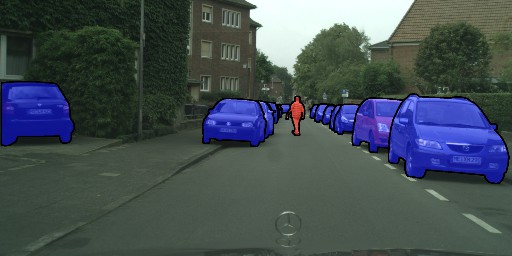}
	\end{subfigure}%
	\hspace{0.03cm}
	\begin{subfigure}{0.33\linewidth}
		\includegraphics[width=0.99\linewidth]{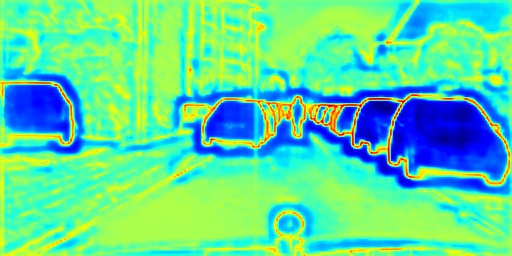}
	\end{subfigure}%
	\hspace{0.01cm}
	\begin{subfigure}{0.33\linewidth}
		\includegraphics[width=0.99\linewidth]{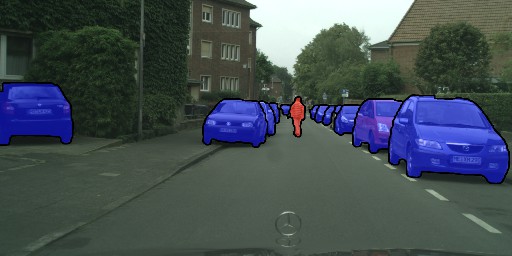}
	\end{subfigure}%
	\\[0.03cm]
	\begin{subfigure}{0.33\linewidth}
		\includegraphics[width=0.99\linewidth]{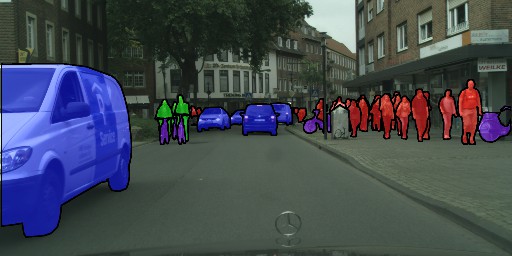}
	\end{subfigure}%
	\hspace{0.01cm}
	\begin{subfigure}{0.33\linewidth}
		\includegraphics[width=0.99\linewidth]{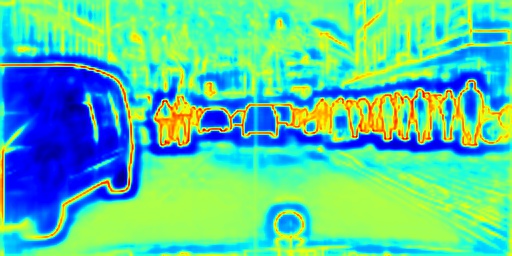}
	\end{subfigure}%
	\hspace{0.01cm}
	\begin{subfigure}{0.33\linewidth}
		\includegraphics[width=0.99\linewidth]{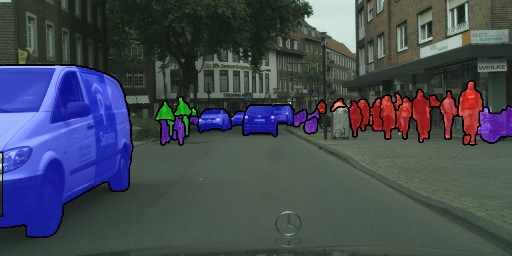}
	\end{subfigure}%
	\\[0.03cm]
	\begin{subfigure}{0.33\linewidth}
		\includegraphics[width=0.99\linewidth]{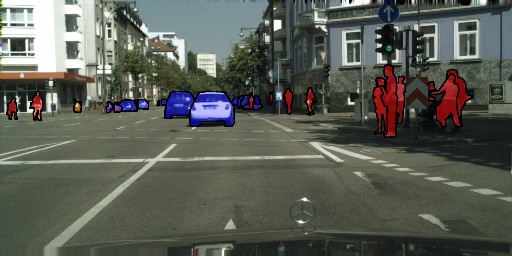}
	\end{subfigure}%
	\hspace{0.01cm}
	\begin{subfigure}{0.33\linewidth}
		\includegraphics[width=0.99\linewidth]{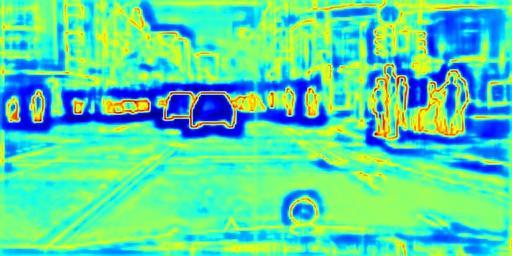}
	\end{subfigure}%
	\hspace{0.01cm}
	\begin{subfigure}{0.33\linewidth}
		\includegraphics[width=0.99\linewidth]{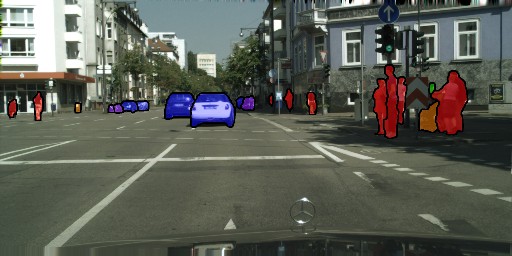}
	\end{subfigure}%
	\\[0.03cm]
	\begin{subfigure}{0.33\linewidth}
		\includegraphics[width=0.99\linewidth]{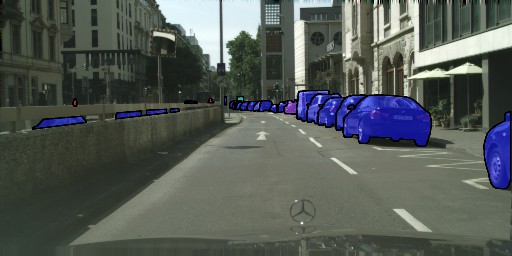}
	\end{subfigure}%
	\hspace{0.01cm}
	\begin{subfigure}{0.33\linewidth}
		\includegraphics[width=0.99\linewidth]{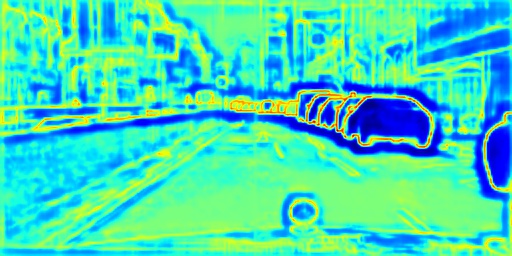}
	\end{subfigure}%
	\hspace{0.01cm}
	\begin{subfigure}{0.33\linewidth}
		\includegraphics[width=0.99\linewidth]{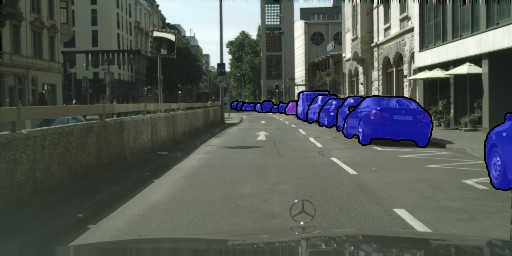}
	\end{subfigure}%
	\\[0.03cm]
	\begin{subfigure}{0.33\linewidth}
		\includegraphics[width=0.99\linewidth]{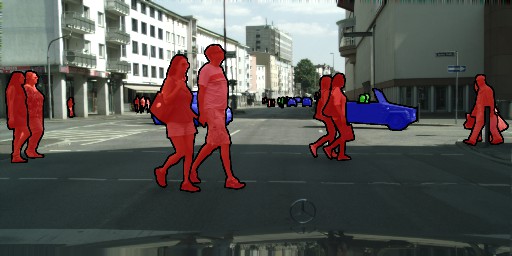}
	\end{subfigure}%
	\hspace{0.01cm}
	\begin{subfigure}{0.33\linewidth}
		\includegraphics[width=0.99\linewidth]{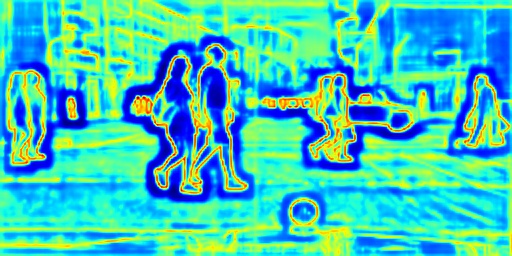}
	\end{subfigure}%
	\hspace{0.01cm}
	\begin{subfigure}{0.33\linewidth}
		\includegraphics[width=0.99\linewidth]{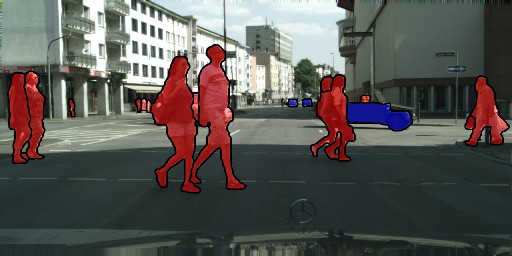}
	\end{subfigure}%
	\\[0.03cm]
	\begin{subfigure}{0.33\linewidth}
		\includegraphics[width=0.99\linewidth]{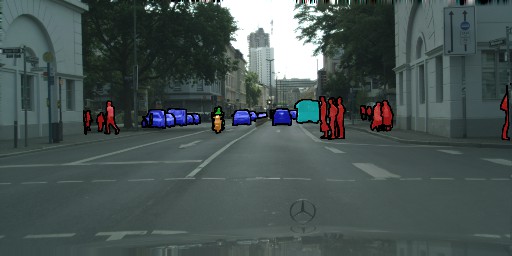}
	\end{subfigure}%
	\hspace{0.01cm}
	\begin{subfigure}{0.33\linewidth}
		\includegraphics[width=0.99\linewidth]{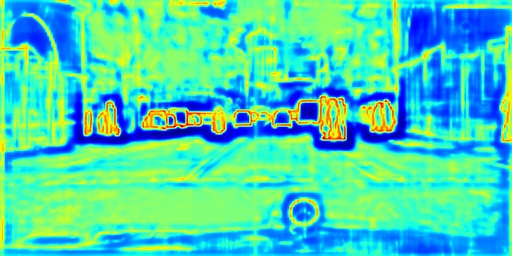}
	\end{subfigure}%
	\hspace{0.01cm}
	\begin{subfigure}{0.33\linewidth}
		\includegraphics[width=0.99\linewidth]{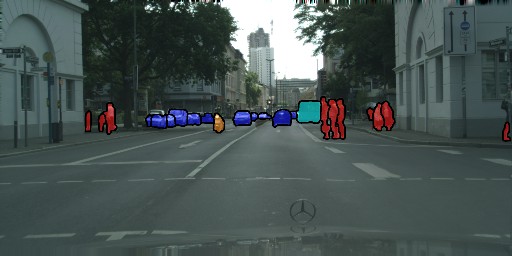}
	\end{subfigure}%
	\\[0.03cm]
	\begin{subfigure}{0.33\linewidth}
		\includegraphics[width=0.99\linewidth]{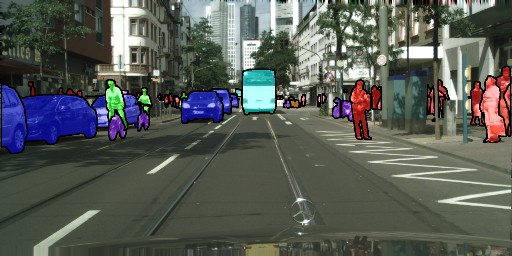}
		\caption{Ground Truth}
	\end{subfigure}%
	\hspace{0.03cm}
	\begin{subfigure}{0.33\linewidth}
		\includegraphics[width=0.99\linewidth]{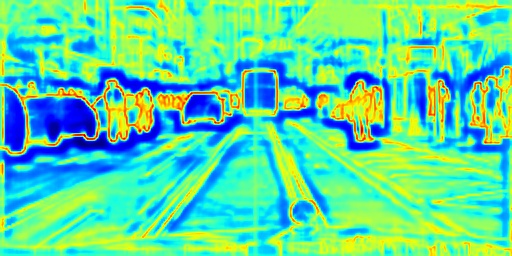}
		\caption{Edges Map}
	\end{subfigure}%
	\hspace{0.03cm}
	\begin{subfigure}{0.33\linewidth}
		\includegraphics[width=0.99\linewidth]{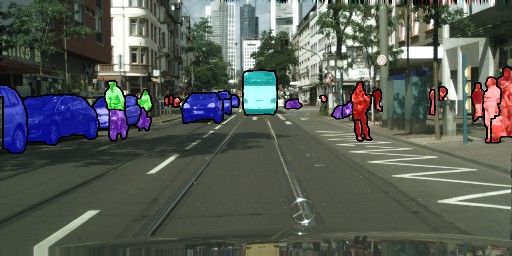}
		\caption{InstanceCut Prediction}
	\end{subfigure}%
	\caption{Curated difficult scene, where InstanceCut performs well. The left column contains input images with ground truth instances highlighted. The middle column depicts per-pixel instance-aware edge log-probabilities and the last column shows the results of our approach.}
	\label{fig:results_curated}
\end{figure*}

\begin{figure*}	
	\begin{subfigure}{0.33\linewidth}
		\includegraphics[width=0.99\linewidth]{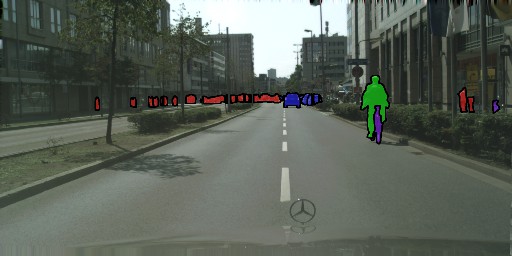}
	\end{subfigure}%
	\hspace{0.03cm}
	\begin{subfigure}{0.33\linewidth}
		\includegraphics[width=0.99\linewidth]{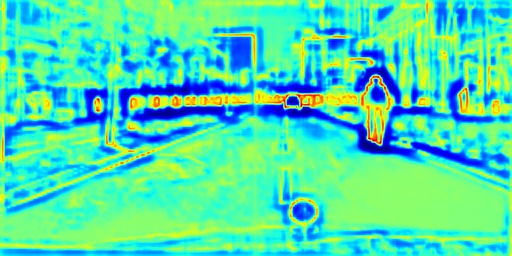}
	\end{subfigure}%
	\hspace{0.01cm}
	\begin{subfigure}{0.33\linewidth}
		\includegraphics[width=0.99\linewidth]{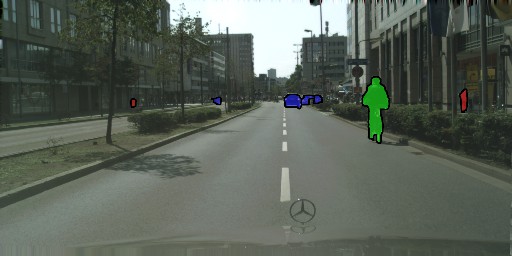}
	\end{subfigure}%
	\\[0.03cm]
	\begin{subfigure}{0.33\linewidth}
		\includegraphics[width=0.99\linewidth]{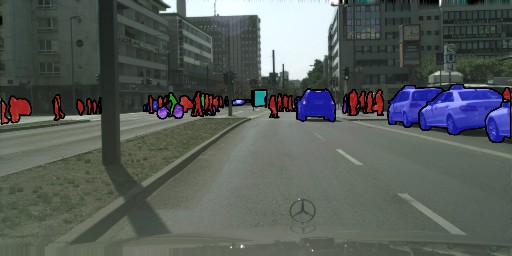}
	\end{subfigure}%
	\hspace{0.01cm}
	\begin{subfigure}{0.33\linewidth}
		\includegraphics[width=0.99\linewidth]{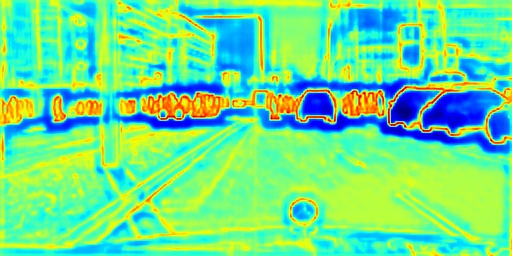}
	\end{subfigure}%
	\hspace{0.01cm}
	\begin{subfigure}{0.33\linewidth}
		\includegraphics[width=0.99\linewidth]{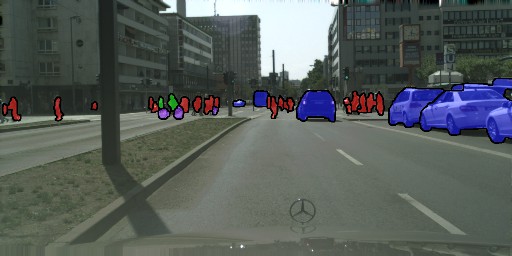}
	\end{subfigure}%
	\\[0.03cm]
	\begin{subfigure}{0.33\linewidth}
		\includegraphics[width=0.99\linewidth]{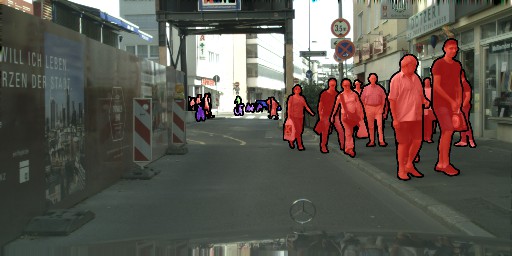}
	\end{subfigure}%
	\hspace{0.01cm}
	\begin{subfigure}{0.33\linewidth}
		\includegraphics[width=0.99\linewidth]{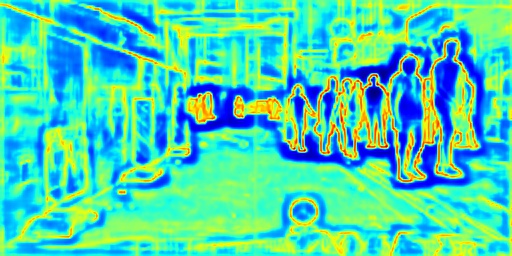}
	\end{subfigure}%
	\hspace{0.01cm}
	\begin{subfigure}{0.33\linewidth}
		\includegraphics[width=0.99\linewidth]{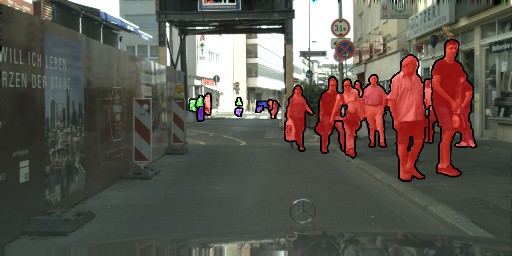}
	\end{subfigure}%
	\\[0.03cm]
	\begin{subfigure}{0.33\linewidth}
		\includegraphics[width=0.99\linewidth]{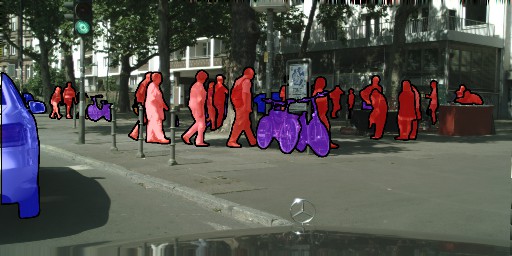}
	\end{subfigure}%
	\hspace{0.01cm}
	\begin{subfigure}{0.33\linewidth}
		\includegraphics[width=0.99\linewidth]{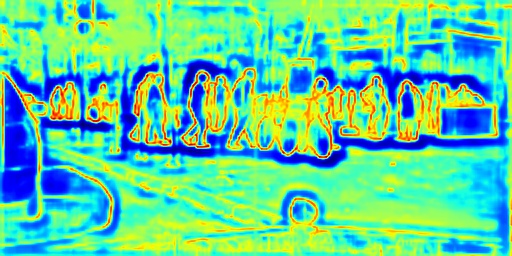}
	\end{subfigure}%
	\hspace{0.01cm}
	\begin{subfigure}{0.33\linewidth}
		\includegraphics[width=0.99\linewidth]{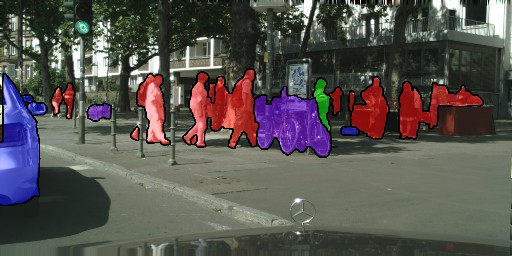}
	\end{subfigure}%
	\\[0.03cm]
	\begin{subfigure}{0.33\linewidth}
		\includegraphics[width=0.99\linewidth]{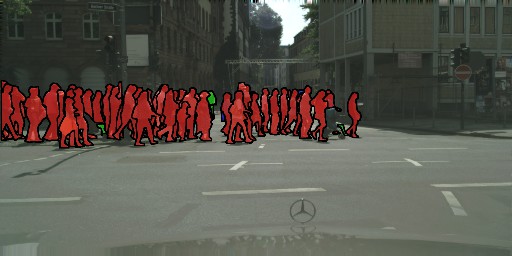}
	\end{subfigure}%
	\hspace{0.01cm}
	\begin{subfigure}{0.33\linewidth}
		\includegraphics[width=0.99\linewidth]{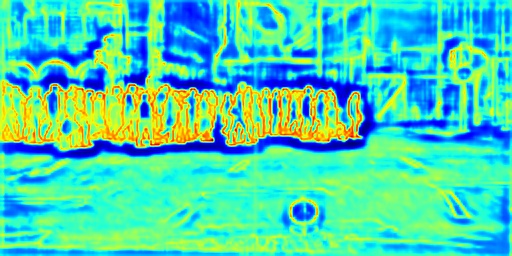}
	\end{subfigure}%
	\hspace{0.01cm}
	\begin{subfigure}{0.33\linewidth}
		\includegraphics[width=0.99\linewidth]{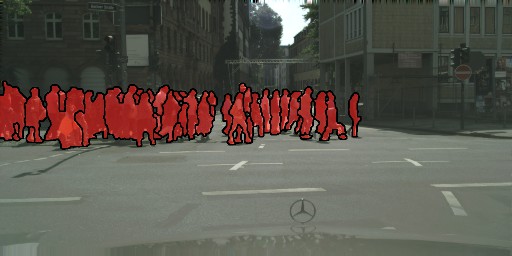}
	\end{subfigure}%
	\\[0.03cm]
	\begin{subfigure}{0.33\linewidth}
		\includegraphics[width=0.99\linewidth]{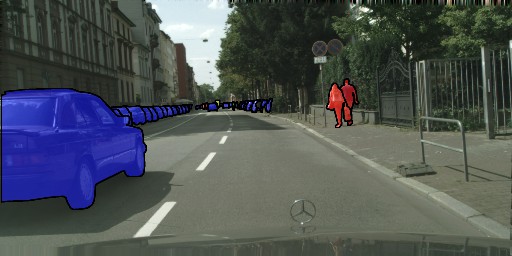}
	\end{subfigure}%
	\hspace{0.01cm}
	\begin{subfigure}{0.33\linewidth}
		\includegraphics[width=0.99\linewidth]{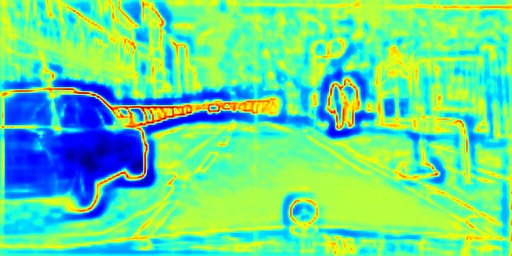}
	\end{subfigure}%
	\hspace{0.01cm}
	\begin{subfigure}{0.33\linewidth}
		\includegraphics[width=0.99\linewidth]{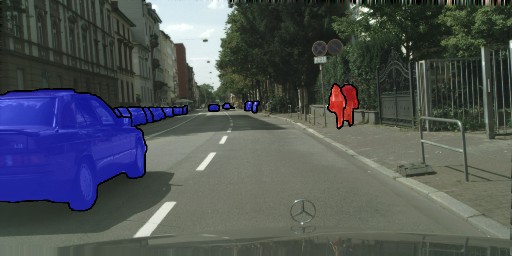}
	\end{subfigure}%
	\\[0.03cm]
	\begin{subfigure}{0.33\linewidth}
		\includegraphics[width=0.99\linewidth]{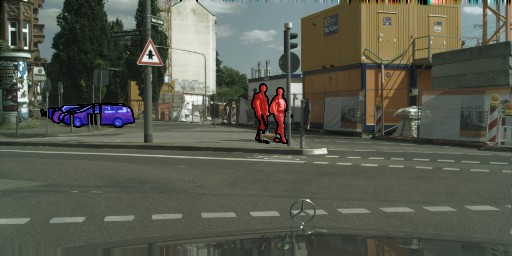}
		\caption{Ground Truth}
	\end{subfigure}%
	\hspace{0.03cm}
	\begin{subfigure}{0.33\linewidth}
		\includegraphics[width=0.99\linewidth]{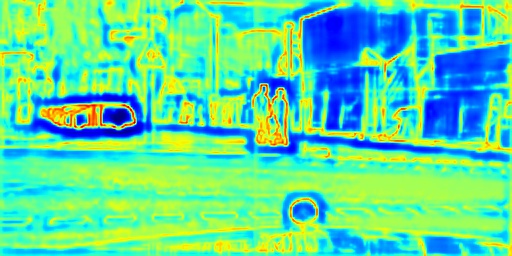}
		\caption{Edges Map}
	\end{subfigure}%
	\hspace{0.03cm}
	\begin{subfigure}{0.33\linewidth}
		\includegraphics[width=0.99\linewidth]{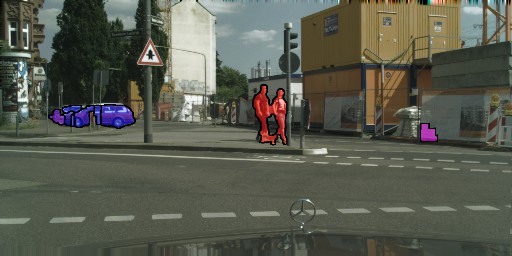}
		\caption{InstanceCut Prediction}
	\end{subfigure}%
	\caption{Failure cases. The left column contains input images with ground truth instances highlighted. The middle column depicts per-pixel instance-aware edge log-probabilities and the last column shows the results of our approach.}
	\label{fig:results_failure}
\end{figure*}

In Table~\ref{tab:results_class} we present a detailed performance comparison. Fig.~\ref{fig:results_curated} contains the subset of difficult scenes where InstanceCut is able to predict most instances correctly. Fig.~\ref{fig:results_failure} contains failure cases of InstanceCut. The main sources of failure are: small objects that are far away from the camera, groups of people that are very close to camera and have  heavy mutual occlusions, and occluded instances that have several disconnected visible parts.

\begin{table}[b]
\begin{minipage}{\textwidth}
\tabcolsep=0.2cm
\center
\begin{tabular}{l|c|c|cccccccc}
Method & Metric & \rot{Mean} & \rot{Person} & \rot{Rider} & \rot{Car} & \rot{Truck} & \rot{Bus} & \rot{Train} & \rot{Motorcycle} & \rot{Bicycle} \\
\toprule
MCG+R-CNN~\cite{Cordts2016Cityscapes} & AP & 4.6 & 1.3 & 0.6 & 10.5 & 6.1 & 9.7 & 5.9 & 1.7 & 0.5 \\
Uhrig et al.~\cite{uhrig2016pixel} & AP & 8.9 & {\bf 12.5} & {\bf 11.7} & 22.5 & 3.3 & 5.9 & 3.2 & 6.9 & {\bf 5.1} \\
InstanceCut & AP & {\bf 13.0} & 10.0 & 8.0  & \textbf{23.7} & \textbf{14.0} & \textbf{19.5} & \textbf{15.2} & {\bf 9.3}  & 4.7  \\
\midrule
MCG+R-CNN~\cite{Cordts2016Cityscapes} & AP$50$\% & 12.9 & 5.6 & 3.9 & 26.0 & 13.8 & 26.3 & 15.8 & 8.6 & 3.1 \\
Uhrig et al.~\cite{uhrig2016pixel} & AP$50$\% & 21.1 & {\bf 31.8} & {\bf 33.8} & 37.8 & 7.6 & 12.0 & 8.5 & 20.5 & {\bf 17.2} \\
InstanceCut & AP$50$\% & {\bf 27.9} & 28.0 & 26.8 & \textbf{44.8} & \textbf{22.2} & \textbf{30.4} & \textbf{30.1} & {\bf 25.1} & 15.7 \\
\midrule
MCG+R-CNN~\cite{Cordts2016Cityscapes} & AP$100$m & 7.7 & 2.6 & 1.1 & 17.5 & 10.6 & 17.4 & 9.2 & 2.6 & 0.9 \\
Uhrig et al.~\cite{uhrig2016pixel} & AP$100$m & 15.3 & {\bf 24.4} & {\bf 20.3} & 36.4 & 5.5 & 10.6 & 5.2 & 10.5 & {\bf 9.2} \\
InstanceCut & AP$100$m & {\bf 22.1} & 19.7 & 14.0 & \textbf{38.9} & \textbf{24.8} & \textbf{34.4} & \textbf{23.1} & {\bf 13.7} & 8.0  \\
\midrule
MCG+R-CNN~\cite{Cordts2016Cityscapes} & AP$50$m & 10.3 & 2.7 & 1.1 & 21.2 & 14.0 & 25.2 & 14.2 & 2.7 & 1.0 \\
Uhrig et al.~\cite{uhrig2016pixel} & AP$50$m & 16.7 & {\bf 25.0} & {\bf 21.0} & 40.7 & 6.7 & 13.5 & 6.4 & 11.2 & {\bf 9.3} \\
InstanceCut & AP$50$m & {\bf 26.1} & 20.1 & 14.6 & \textbf{42.5} & \textbf{32.3} & \textbf{44.7} & \textbf{31.7} & {\bf 14.3} & 8.2 \\
\bottomrule
\end{tabular}
\caption{CityScapes results. Instance-aware semantic segmentation results on the test set of CityScapes, given for each semantic class. The table contains only published results.}
\label{tab:results_class}
\end{minipage}
\end{table}

\end{document}